\pdfoutput=1

\documentclass[11pt]{article}

\usepackage[final]{acl}

\usepackage{times}
\usepackage{latexsym}

\usepackage[T1]{fontenc}

\usepackage[utf8]{inputenc}

\usepackage{microtype}

\usepackage{inconsolata}
\usepackage{booktabs}
\usepackage{graphicx}

\title{Visualising Policy-Reward Interplay to Inform Zeroth-Order Preference Optimisation of Large Language Models}

\author{Alessio Galatolo, Zhenbang Dai, Katie Winkle, Meriem Beloucif \\
       Uppsala University\\
    \texttt{\{alessio.galatolo, katie.winkle\}@it.uu.se}\\
    \texttt{zhenbang.dai.8990@student.uu.se}\\
    \texttt{meriem.beloucif@lingfil.uu.se}
}
\usepackage{amsmath, amsfonts}
\usepackage{algorithm}
\usepackage{algpseudocode}
\usepackage{subcaption}
\usepackage{listings} 
\usepackage{adjustbox}

\begin{document}
\maketitle
\begin{abstract}
Fine-tuning Large Language Models (LLMs) with first-order methods like back-propagation is computationally intensive. Zeroth-Order (ZO) optimisation uses function evaluations instead of gradients, reducing memory usage, but suffers from slow convergence in high-dimensional models. As a result, ZO research in LLMs has mostly focused on classification, overlooking more complex generative tasks. In this paper, we introduce ZOPrO, a novel ZO algorithm designed for \textit{Preference Optimisation} in LLMs. We begin by analysing the interplay between policy and reward models during traditional (first-order) Preference Optimisation, uncovering patterns in their relative updates. Guided by these insights, we adapt Simultaneous Perturbation Stochastic Approximation (SPSA) with a targeted sampling strategy to accelerate convergence. Through experiments on summarisation, machine translation, and conversational assistants, we demonstrate that our method consistently enhances reward signals while achieving convergence times comparable to first-order methods. While it falls short of some state-of-the-art methods, our work is the first to apply Zeroth-Order methods to Preference Optimisation in LLMs, going beyond classification tasks and paving the way for a largely unexplored research direction. Code and visualisations are available at \url{https://github.com/alessioGalatolo/VisZOPrO}.
\end{abstract}

\section{Introduction}
Fine-tuning large language models (LLMs) is a computationally demanding process, often incurring significant costs in terms of both time and memory. Traditional fine-tuning relies on first-order optimisation techniques, such as back-propagation \cite{rumelhart1986learning}, which require substantial computational resources due to the calculation and storage of gradients. The memory footprint for storing model parameters, activations, and gradients can be prohibitive, especially for large-scale LLMs \cite{ 10.1145/3442188.3445922, doi:10.1021/acs.est.3c01106, Rimban_2023, 10765824}.

\begin{figure}[t]
    \centering
    \includegraphics[width=\columnwidth]{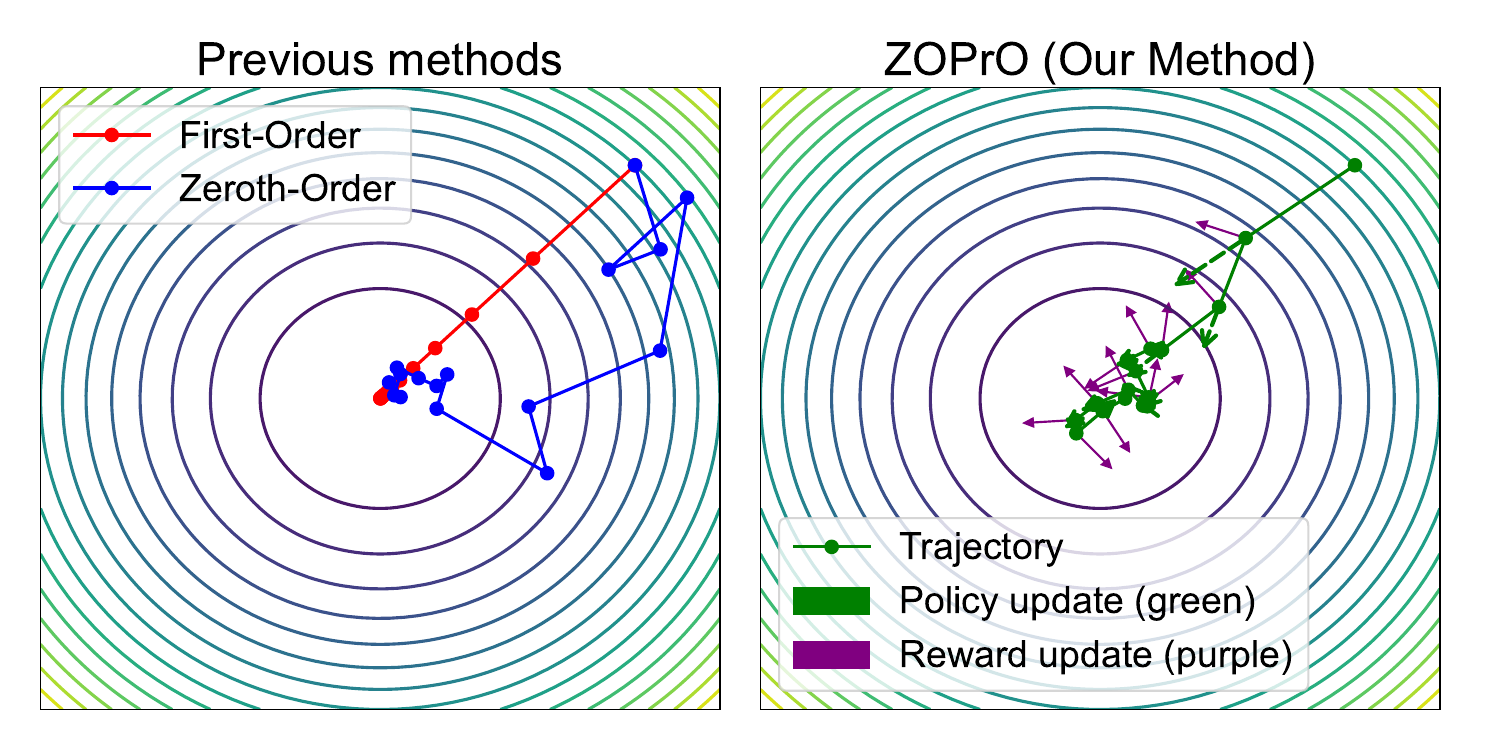}
    \caption{Illustration of how First and Zeroth-Order methods would progress in the optimisation space (left) and our method's informed progress (right).}
    \label{fig:zopro}
\end{figure}

This has spurred interest in alternative optimisation techniques, such as Zeroth-Order Optimisation \cite{mezo, zhang2024revisiting, shirkavand2025bilevelzofobridgingparameterefficient}. Zeroth-Order optimisation (Figure \ref{fig:zopro}, left) presents a compelling alternative by eliminating the need for gradient computation. Instead, it approximates gradients using only function evaluations (i.e., forward passes), typically through finite difference methods. For example, in Simultaneous Perturbation Stochastic Approximation (SPSA) \cite{spall1992multivariate, mezo}, the function is evaluated two times, where each time noise is either added or subtracted from the weights and a `step' is taken toward the most promising direction. This significantly reduces memory requirements, as the model needs to be loaded into memory only once. Additionally, due to their very nature, these methods are also inherently fit for optimising non-differentiable objectives such as human preferences \cite{mezo} where a step can be taken directly towards the preferred model's generations. While promising, Zeroth-Order methods face challenges in terms of convergence speed, particularly for high-dimensional models like LLMs.

Recent research has explored various strategies to improve the convergence of Zeroth-Order methods for LLM fine-tuning. These include using strong fixed prompts \cite{mezo, zhang2024revisiting}, employing meta-learning techniques and combining zeroth-order with first-order methods \cite{shirkavand2025bilevelzofobridgingparameterefficient}. However, all empirical studies to date focus on classification or question-answering tasks, which may not fully capture the complexities associated with fine-tuning LLMs for more complex generative tasks.

In contrast to previous work, we dive into the more challenging and practically relevant problem of Preference Optimisation for LLMs. Preference optimisation involves training a policy model to generate desirable outputs, often guided by a reward model that learns to distinguish between high-quality and low-quality outputs. This process, in many cases, entails iterative training of both models, generally in settings of online learning \cite{onlineDPO} or as a way to refine the reward model to account for the distribution shift of the policy throughout training \cite{wang2024secretsrlhflargelanguage}.

To this end, we hypothesise that the evolving dynamics of the policy and reward models during Preference Optimisation can provide valuable information for guiding the search direction of Zeroth-Order methods. We thus begin our investigation by visualising and analysing how popular preference optimisation methods, such as Proximal Policy Optimization (PPO) \cite{schulman2017proximalpolicyoptimizationalgorithms, rlhf} and Direct Preference Optimization\footnote{As standard DPO does not use any reward directly, we choose one of its variations \cite{onlineDPO} where we can integrate the reward model.} (DPO) \cite{dpo}, act on the policy and reward model, with particular attention on their relation. Whilst we do not find any statistical \textit{direct} correlation, our visualisation and analysis suggest a significant influence of previous updates on upcoming ones. Further, we find that the reward model's updates are consistently orthogonal to those of the policy.

We integrate these findings into traditional SPSA by influencing how the random vector is sampled, restricting the sample space to improve the convergence rate. We then show how our modifications do not hinder the theoretical guarantee of convergence. Finally, we evaluate our method (sketched in Figure \ref{fig:zopro}) on various tasks and datasets to show its effectiveness.

Our contributions are as follows:

\begin{enumerate}
    \item We employ dimensionality reduction techniques to visualise the evolution of policy and reward model parameters during preference optimisation, providing insights into the underlying dynamics of the training process.
    \item Based on the observed dynamics, we develop ZOPrO, a Zeroth-Order Optimisation algorithm specifically tailored for preference optimisation in LLMs. ZOPrO incorporates a novel mechanism to guide the search direction by utilising the inter-dependencies between the policy and reward models.
    \item We conduct experiments on multiple tasks and datasets to demonstrate the effectiveness of ZOPrO in fine-tuning LLMs for Preference Optimisation. Our results show that ZOPrO achieves comparable performance to first-order methods while significantly reducing memory usage and iteration time.
\end{enumerate}

\section{Related Works}

\subsection{Preference Optimisation}
Reinforcement Learning from Human Feedback (RLHF), or more generally, Preference Optimisation, aims to optimise Machine Learning models to align with complex human judgements and preferences \cite{christiano2017deep}. Typically, human feedback data is used to train a reward model, often conceptualised with the Bradley-Terry model \cite{bradley1952rank}, as an effective approach to converting human preferences into continuous reward signals \citep{bai2022training}. The reward model is then used to optimise a policy model using Reinforcement Learning algorithms, such as Proximal Policy Optimization (PPO) \cite{schulman2017proximalpolicyoptimizationalgorithms}. RLHF has been applied in LLMs across various tasks from sentiment generation \cite{ziegler2019fine} to summarisation \cite{stiennon2020learning} to instruction following \cite{rlhf}. In this framework, the policy and reward models influence each other in a shared interplay. Updates to the policy shift the text distribution that the reward model evaluates \citep{lu2024takes,wang2024secretsrlhflargelanguage, dou2024metarm}, while updates to the reward model alter the policy's optimisation trajectory \citep{li-etal-2024-optimizing-language}.

Newer methods, such as Direct Preference Optimization (DPO) \cite{dpo} and variations \cite{liu2024statistical, pmlr-v238-gheshlaghi-azar24a, ji2024towards} avoid the use of Reinforcement Learning, instead relying on \textit{indirect} reward. However, other works suggest that the use of a reward model or an external judge can still be beneficial \cite{onlineDPO}. Whilst promising, these methods have not replaced RL-based approaches. Newer RL approaches include Group Relative Policy Optimization (GRPO) \cite{shao2024deepseekmath}, which introduces group-based relative rewards to accelerate convergence and obtain substantial improvements in math and general reasoning \cite{deepseekai2025deepseekr1incentivizingreasoningcapability} or \cite{ahmadian2024back} which introduces REINFORCE Leave-One-Out (RLOO), a REINFORCE-based approach competitive to PPO with gains in memory utilisation and convergence speed.

Recent works have also explored operations in the weight space to enhance preference alignment. ExPO \cite{zheng2025expo} proposes a linear extrapolation technique to improve model performance after partial or full preference optimisation. Related trends include Rewarded Soups \cite{rame2023rewarded}, which uses Linear Mode Connectivity (LMC) \cite{frankle2020lmc} to interpolate between models optimised on diverse reward signals. These methods rely on the assumption that preference-optimised models remain connected via simple geometric transformations, such as \textit{linear} paths in parameter space. While compelling, this assumption is not general and may not (and as we will show, does not) hold in settings where policy and reward models evolve under distinct and potentially divergent optimisation pressures.

\subsection{Visualising LLMs}
While visualisation techniques are used to understand various aspects of LLMs, few studies directly visualise the model weights.  Many focus on visualising the loss landscape to understand generalisation \cite{yu2024superweightlargelanguage}, or use dimensionality reduction techniques like t-SNE \cite{tsne} to visualise the distribution of samples in the LLM's representation space \cite{yuan2024llminferenceunveiledsurvey}. Others explore attention mechanisms and how the model attends to different tokens \cite{vig-2019-multiscale}, or visualise code generation as an optimisation problem \cite{light2024scatteredforestsearchsmarter}.

No works explore how the weights move in the parameter space, and no work specifically focuses on the interplay between reward and policy models in preference optimisation.

\subsection{Zeroth-Order Methods}
While Zeroth-Order (ZO) Optimisation methods have existed for decades \cite{spall1992multivariate}, their application to large language models (LLMs) is relatively recent \cite{zhang2024revisiting}. This resurgence is driven by the need for memory-efficient fine-tuning techniques \cite{mezo}, particularly for on-device personalisation \cite{guo2024zerothorder} and addressing the challenges of ever-growing model sizes. Many works \cite{shirkavand2025bilevelzofobridgingparameterefficient, yu2024zerothorderfinetuningllmsrandom, chen2024enhancingzerothorderfinetuninglanguage, jiang2024zo} have since applied Zeroth-Order to LLMs, however, all of them explore only classification or question-answering tasks, questioning the method's applicability to more complex generative tasks such as Preference Optimisation. Works such as \cite{zhang2025zerothorder} provide \textit{theoretical} insights into the combination of ZO and Reinforcement Learning from Human Feedback (RLHF), however, their proposed methods are impractical in real-world scenarios, as they suffer from extreme convergence issues, rendering them infeasible.

\section{Visualising Preference Optimisation}
\label{sec:vis}
We begin our investigation by analysing how the policy and reward models are optimised using traditional methods. We simulate multiple training runs across different methods, collecting model checkpoints at various stages of optimisation. At the end of training, we visualise and analyse the weights as updated by each method with particular attention to the interplay between the reward and policy models.

\subsection{Experimental Setup}
\label{sec:vis_setup}
To balance computational efficiency and model performance, we conduct our experiments on Llama-3.2 1B \cite{llama3}. This model size enables us to complete training with relatively low resource consumption while still maintaining the capability to be optimised for preference learning, as Llama's suite already provides a tuned version of such a model. 

We select the two most used methods for preference optimisation: PPO \cite{schulman2017proximalpolicyoptimizationalgorithms} and DPO \cite{dpo}; for the latter, we select its \textit{online} version \cite{onlineDPO} which integrates a reward model---an essential component for our analysis. The models are trained using Anthropic's HH-RLHF dataset \cite{bai2022training}, chosen due to its widespread adoption in preference modelling tasks. After initial testing, we report convergence difficulties with PPO, where the model easily collapses into generating only eos tokens. Due to this, we instead analyse one of its variations, RLOO \cite{ahmadian2024back}. RLOO is very similar to PPO in many aspects, but differs in how it treats actions and in the complexity of its objective. PPO treats each generated token as an individual action. It computes advantages with an extra value model\footnote{Here note that even for PPO, we use the same model both as the value and reward for simplicity.} using Generalized Advantage Estimation (GAE), and optimises a clipped policy ratio. RLOO simplifies this by treating the entire completion as a single action, avoiding the value model altogether. It uses a REINFORCE-style objective with a leave-one-out baseline, sampling 
$k$ ($=2$ in our setting) rollouts per query.

For all methods, we adopt an iterative training regime in which, after each iteration, the reward model is updated to better evaluate samples from the policy model, accounting for distribution shifts. We show pseudocode for our training regime in Algorithm \ref{alg:iterative_optimization}. For the reward, we use the standard Bradley-Terry model \cite{btmodel}.

\begin{algorithm*}[ht]
  \caption{Iterative Policy Optimisation and Reward Model Refinement}
  \label{alg:iterative_optimization}
  \begin{algorithmic}[1]
    \Require SFT model $\pi_{SFT}$, Reward model $R_0$, Number of iterations $N$, Prompt dataset $D_{prompt}$

    \State $\pi_0 \gets \pi_{SFT}$  \Comment{Initialize policy}

    \For{$i \gets 1$ to $N$}
      \Comment{1. Policy Optimisation (One Iteration)}
      \State $\pi_i \gets \text{OptimisePolicy}(\pi_{i-1}, R, \pi_{SFT}, D_{prompt})$ \Comment{Optimize policy using current reward model}

      \Comment{2. Data Collection to update reward model}
      \State $dataset \gets []$ \Comment{Initialize dataset for reward model update (list of tuples)}
      \For{each prompt $p$ in  $D_{prompt}$}
        \State $gen_1 \gets \text{Generate}(\pi_i, p)$ \Comment{Generate with current policy}
        \State $gen_2 \gets \text{Generate}(\pi_i, p)$ \Comment{Generate a second response for the same prompt}
        \State $accepted, rejected \gets \text{Judge}(gen_1, gen_2, p)$ \Comment{Get external judge's preference}
        \State $dataset.append((p, accepted, rejected))$ \Comment{Add tuple to dataset}
      \EndFor

      \Comment{3. Reward Model Update, adjusting for distribution shift}
      \State $R_i \gets \text{RefineRewardModel}(R_{i-1}, dataset)$ \Comment{Update reward model using the new data}

    \EndFor

    \Return [$\pi_1$, ..., $\pi_N$], [$R_1$, ..., $R_N$]
  \end{algorithmic}
\end{algorithm*}

Tracking weight updates at each iteration is essential for our analysis but poses challenges in terms of storage and computational costs. To mitigate this, we train using Low-Rank Adapters (LoRA) \cite{hu2022lora}, which offers a compact, low-rank representation of the model's weights while preserving optimisation dynamics. We save one checkpoint per iteration and additional intra-iteration snapshots. 

\subsection{Visualisation and Analysis}
Once the checkpoints are collected, we analyse them using popular dimensionality reduction techniques: t-SNE \cite{tsne} and UMAP \cite{umap}. For t-SNE we use openTSNE \cite{Policar2024}, with FIt-SNE's \cite{linderman2019fast} implementation. We use these methods to project the model's weights into 2D and 3D spaces for visualisation.

UMAP is particularly useful due to its ability to preserve global structure, while t-SNE excels in capturing local neighbourhoods. Given that Preference Optimisation involves structured learning dynamics, we prioritise UMAP for global trajectory analysis. To enable further \textit{quantitative} analysis, we utilise reduced dimensional data (to 10000 using UMAP), retaining key structural patterns while making computations feasible.

\subsection{Results}
\begin{figure*}
    \centering
    \begin{subfigure}[b]{0.24\textwidth}
        \centering
        \includegraphics[width=\textwidth]{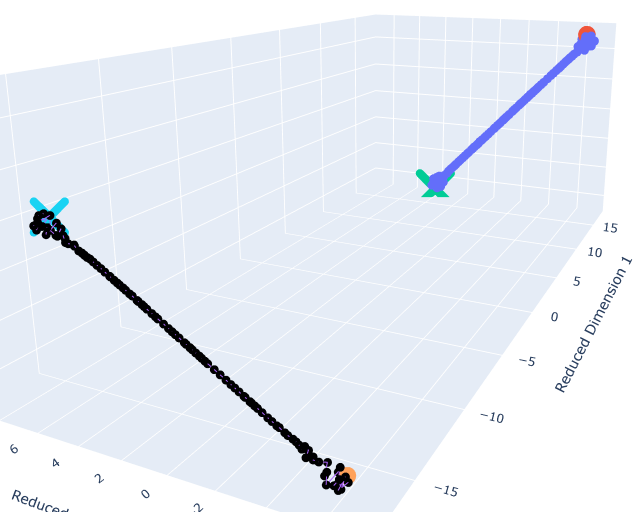}
        \caption{Online DPO, t-SNE}
    \end{subfigure}
    \begin{subfigure}[b]{0.24\textwidth}
        \centering
        \includegraphics[width=\textwidth]{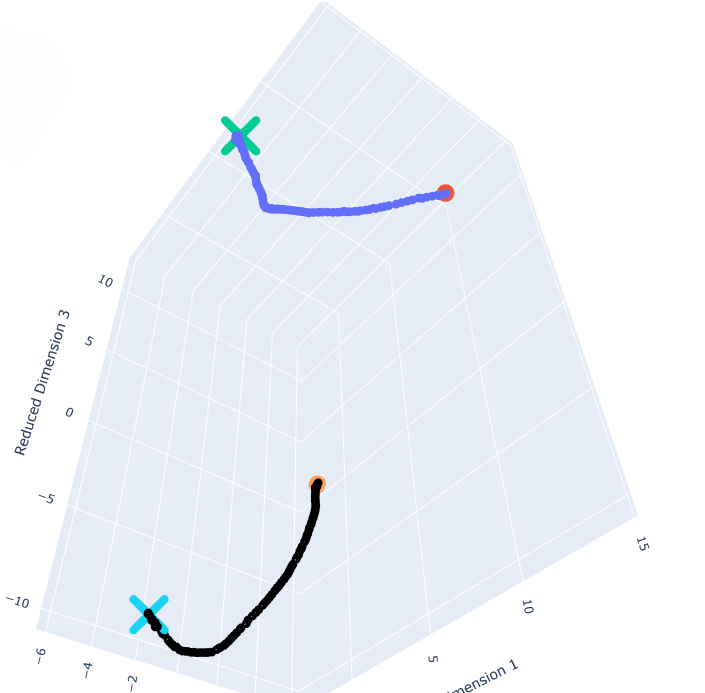}
        \caption{Online DPO, UMAP}
    \end{subfigure}
    \begin{subfigure}[b]{0.24\textwidth}
        \centering
        \includegraphics[width=\textwidth]{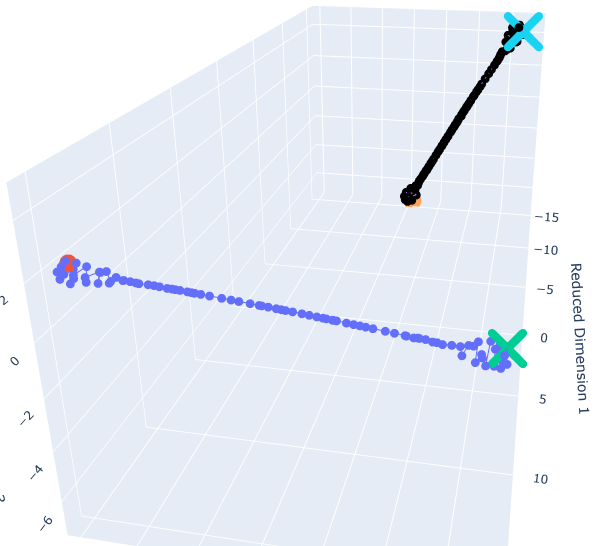}
        \caption{RLOO, t-SNE}
    \end{subfigure}
    \begin{subfigure}[b]{0.24\textwidth}
        \centering
        \includegraphics[width=\textwidth]{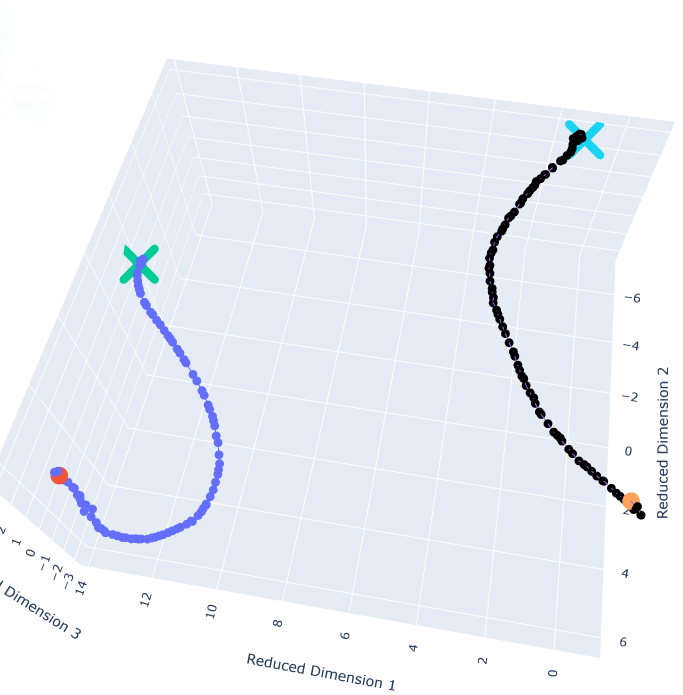}
        \caption{RLOO, UMAP}
    \end{subfigure}
    \caption{3D visualisation of training. Policy model is in black, reward is in blue and the cross indicates the last checkpoint.}
    \label{fig:visuals}

\end{figure*}

\begin{figure}
    \centering
    \begin{subfigure}[b]{0.23\textwidth}
        \centering
        \includegraphics[width=\textwidth]{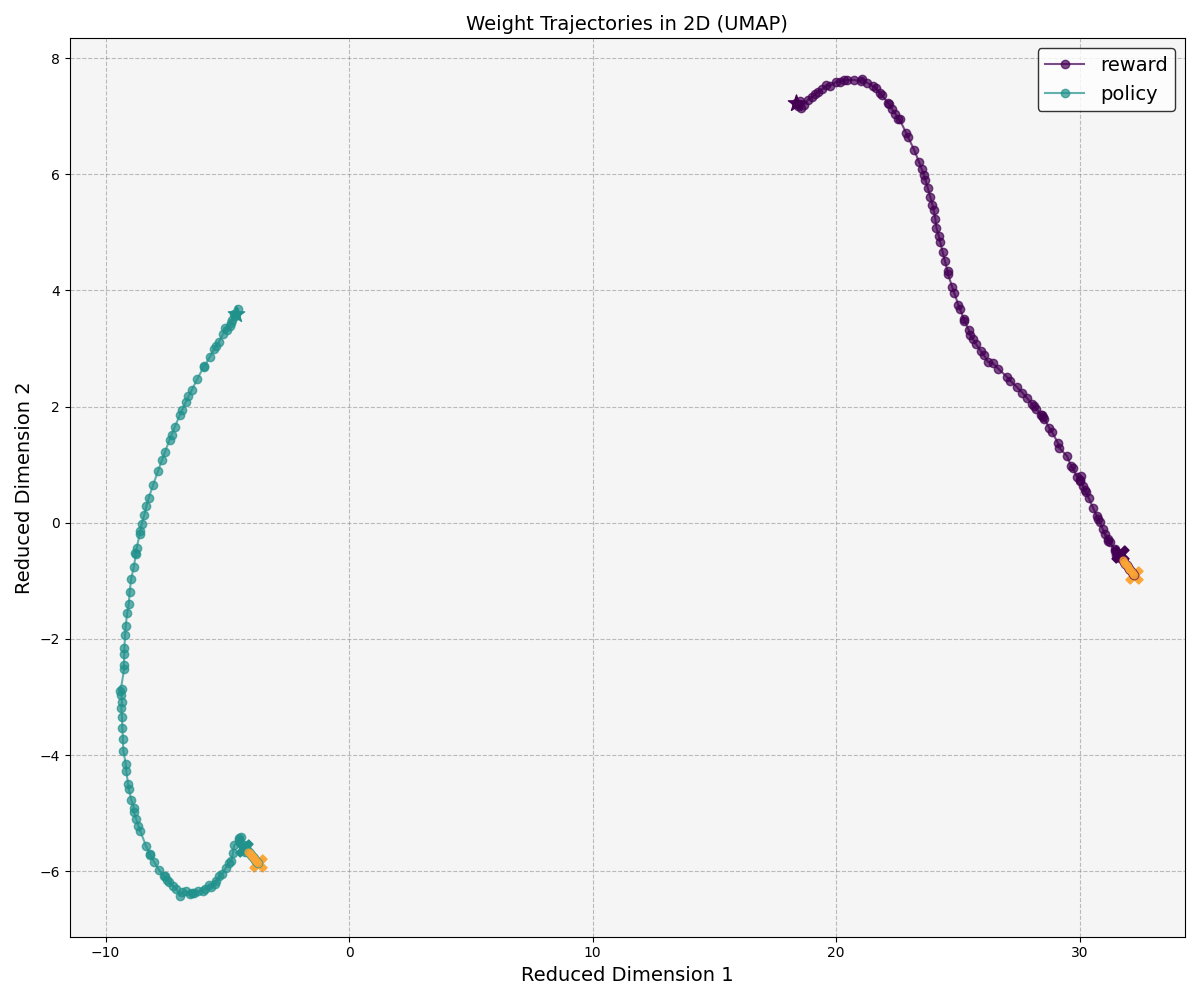}
        \caption{Online DPO, UMAP}
    \end{subfigure}
    \begin{subfigure}[b]{0.23\textwidth}
        \centering
        \includegraphics[width=\textwidth]{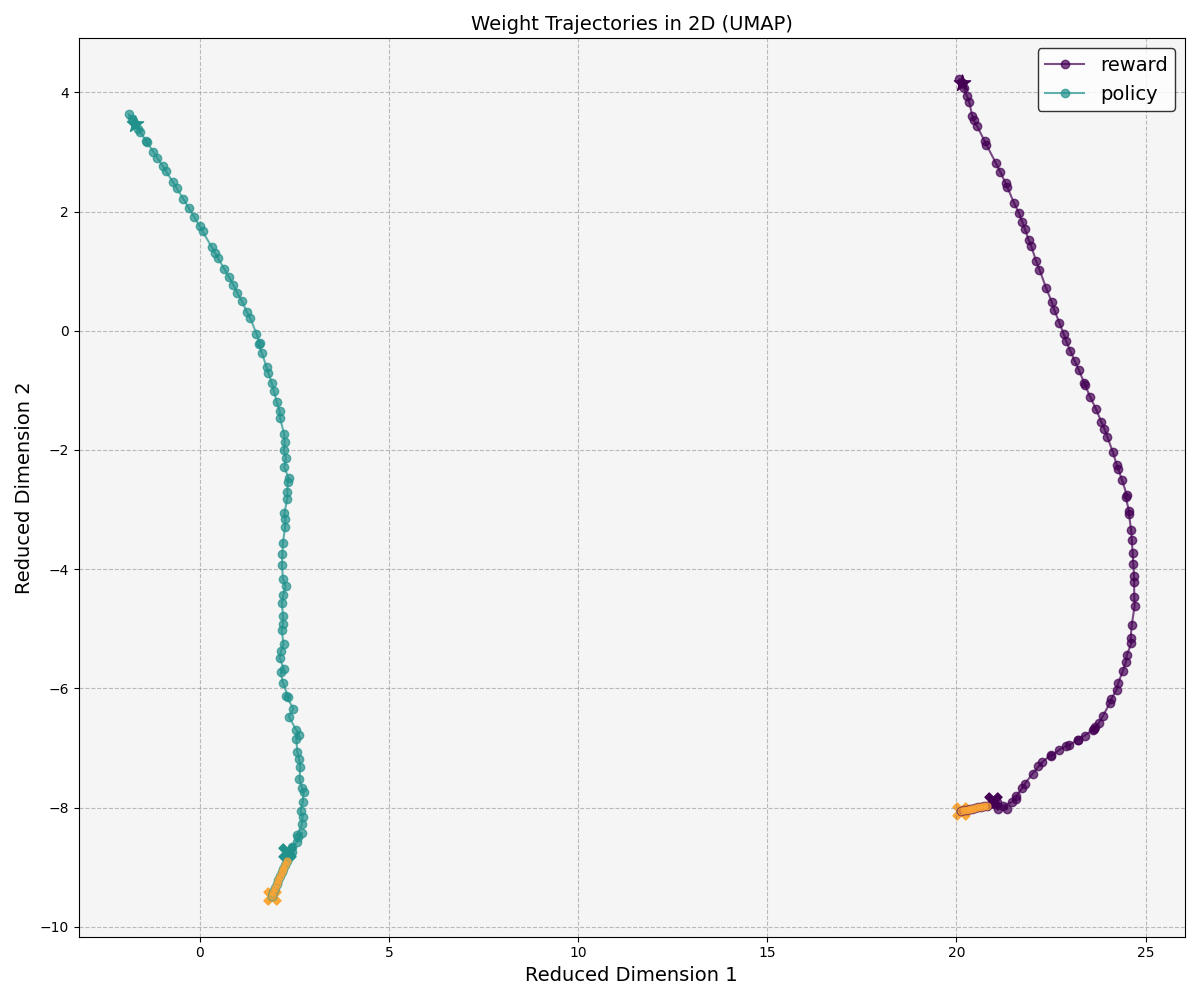}
        \caption{RLOO, UMAP}
    \end{subfigure}
    \caption{3D visualisation of training. Policy is in cyan, while in purple is the reward model. In orange we project a prediction of upcoming steps.}
    \label{fig:2d-visuals}
\end{figure}

We show some 2D visualisations of different methods in Figure \ref{fig:2d-visuals} and 3D visualisations in \ref{fig:visuals}. We include interactive visualisations at \url{https://alessiogalatolo.github.io/VisZOPrO/}.

Figure \ref{fig:visuals} presents a comparison of different preference optimisation methods by tracking the trajectory of both the policy and reward models throughout training. Each subfigure illustrates a different optimisation method, with t-SNE and UMAP used to visualise the evolution of the models' weights. The policy model's trajectory is depicted in black, while the reward model's trajectory is in blue. The final checkpoint is marked with a cross.

From the visualisation, we observe distinct convergence behaviours across methods. Online DPO (Figures \ref{fig:visuals}a and \ref{fig:visuals}b) shows a structured trajectory with a relatively stable alignment between the policy and reward models. In contrast, RLOO (Figures \ref{fig:visuals}c and \ref{fig:visuals}d) exhibits a more erratic pattern, indicating a less stable optimisation process. Furthermore, UMAP projections (Figures \ref{fig:visuals}b and \ref{fig:visuals}d) seem to capture global structure better than t-SNE, which tends to emphasise local neighbourhoods.

These insights align with existing literature, where DPO has been shown to exhibit more stable training dynamics compared to RL-based approaches. On the other hand, the non-linear trajectory of optimisation goes in contrast with some literature on model merging \cite{zheng2025expo}, which attempts to obtain fully trained models by \textit{linear} extrapolation of partial training weights. 

\subsection{Quantitative Analysis}
To complement our visual analysis, we conduct a quantitative evaluation using several key metrics. Unless otherwise specified, all analyses were performed on dimensionally reduced data (to 10000 dimensions via UMAP) due to the prohibitive memory requirements of operating in the original weight space. We report here the most important finding but we also include more details and plots in the Appendix.

Starting from (online) DPO, we record a Procrustes disparity between the reward and policy models of 0.767, indicating a notable but not extreme divergence between their respective trajectories. This suggests that while the models undergo different updates, their overall evolution maintains a degree of coherence.

Additionally, we track the distance correlation between the reward and policy models across training iterations. While initial correlation is relatively low (0.180 at time step 0), it exhibits fluctuations throughout training, with notable peaks of 0.241, 0.313 and 0.253. This suggests that the interaction between the models is not constant but undergoes periodic alignment and divergence.

To further examine alignment, we compute the relative angle between weight vectors at each step. The observed angles are mostly around 90$\pm$5 degrees with small oscillations in certain stages. 

Finally, we compute the Pearson correlation of weight changes between reward and policy models. The values fluctuate significantly, with both positive and negative correlations observed throughout training, reinforcing the hypothesis that the optimisation trajectories are non-monotonic but exhibit periodic phases of alignment.

For RLOO, the Procrustes disparity is higher at 0.832, suggesting greater divergence between the reward and policy models. Distance correlation remains more variable and lower overall, with values ranging from 0.141 to 0.262. The relative angles between weight vectors show similar trends, with fewer instances of oscillation compared to Online DPO. Pearson correlation of weight changes further highlights this instability, with large fluctuations between positive and negative correlations throughout training.

These findings indicate that Online DPO maintains a structured but dynamic interaction between reward and policy models, contributing to its observed stability. In contrast, RLOO demonstrates higher divergence and weaker alignment, consistent with its more erratic optimisation trajectory observed in the visualisations.

Motivated by our observations on relative angles, we conduct a detailed analysis of orthogonality. Specifically, we examine model weights in their original dimensional space, assessing orthogonality both at the overall model level and layer-wise. Our empirical findings demonstrate consistent orthogonality between reward and policy updates across both methods. While both models originate from the same pretrained model (e.g., Llama), their updates quickly diverge due to distinct optimisation objectives. At iteration 0, this scenario closely mirrors that of \citet{yu2020gradient}, who showed that conflicting gradient updates naturally arise in multi-objective contexts with shared initial parameters. As training progresses, however, the analogy weakens due to increasing parameter divergence. Nonetheless, our results extend the broader insight empirically into the context of two interacting but distinct processes---reward modelling, which employs discriminative supervised learning, and policy optimisation, which involves generative preference optimisation.

\section{ZOPrO: Zeroth-Order Preference Optimisation}
In this section, we introduce Zeroth-Order Preference Optimisation (ZOPrO), a method for optimising policies in RLHF-like settings without explicit gradient information. We first formalise the iterative training process, describing our adaptation of Simultaneous Perturbation Stochastic Approximation (SPSA) for policy optimisation, and finally present our novel sampling strategy that takes into account the insights gained with our visualisation analysis, in particular, we exploit the orthogonality between policy and reward updates and the relation between each policy update and its consequent. \footnote{Our code is available at \url{https://github.com/alessioGalatolo/VisZOPrO}, we also include training details in the Appendix to favour reproducibility.}

\subsection{Formalisation}
We consider a policy model \(\pi_\theta\) trained iteratively using a reward model \(R_\phi\). The reward model is updated at the end of each iteration to account for the distribution shift induced by changes in \(\pi_\theta\). Here, \(\theta \in \mathbb{R}^d\) represents the parameters of the policy model, while \(\phi \in \mathbb{R}^{d+b}\) includes both the shared parameters and the additional value head of \(R_\phi\) with \(b\) extra parameters.

Let \(\Delta\pi_t\) denote the difference between consecutive policy iterations \(\pi_\theta^{(t)}\) and \(\pi_\theta^{(t-1)}\), and define \(\Delta R_t = R_\phi^{(t)} - R_\phi^{(t-1)}\). For ease of notation, we omit explicit dependencies on \(\theta\) and \(\phi\), focusing only on shared parameters (thus excluding \(R\)'s value head). Each training iteration can consists of multiple steps and epochs, forming a complete training cycle. Our empirical findings suggest that \(\Delta\pi_{t+1}\) exhibits high similarity to \(\Delta\pi_t\) and remains largely orthogonal to \(\Delta R_t\).

Our proposed ZOPrO method utilises this structure in conjunction with SPSA \cite{spall1992multivariate}, as the base of the ZO algorithm, and RLOO \cite{ahmadian2024back}, as the objective function. \footnote{Whilst we use RLOO as an exemplar objective, our method can be easily extended to other RL-based objectives without substantial changes.}

\subsubsection{Optimisation via SPSA}
SPSA is a gradient-free optimisation algorithm that estimates gradients using function evaluations at perturbed parameter values:
\begin{equation}
    \widehat{g}_i \approx \frac{\mathcal{J}(\theta_i + \varepsilon_i z_i) - \mathcal{J}(\theta_i - \varepsilon_i z_i)}{2 \varepsilon_i} z_i,
\end{equation}
where \(z_i\) is a random perturbation direction (e.g., Gaussian or Rademacher noise), and \(i\) denotes an individual step within iteration \(t\). Under standard smoothness assumptions, this scheme ensures convergence to a local optimum given a suitable step-size schedule.

In our approach, we retain the RLOO objective function \(\mathcal{J}(\theta)\) while modifying how \(z_i\) is sampled to enhance convergence. Specifically, we reduce the sampling space by incorporating knowledge from previous updates.

Note how our approach is also validated by previous attempts at surrogate or prior-guided strategies in ZO methods such as \cite{cheng2021on, meier2019improvinggradientestimationevolutionary}, which influence their sampling through the use of multiple `oracle gradient directions'. In our case, we ground our perturbation strategy in empirical evidence and reuse the updates from the previous iteration. In particular, prior works had to use `oracle' directions as ZO updates are very noisy locally and cannot be used directly. Our method instead leverages the weight updates over a whole iteration, making their use possible.

\subsubsection{Adaptive Perturbation Strategy}
To compute perturbation directions \(z\), we first utilise \(\Delta\pi_t\) with a scaling factor \(\alpha\), which starts at 1 and decays to 0 throughout each iteration. Next, we sample a random noise vector from a multivariate standard normal distribution, \(\mathcal{N}(0, I_d)\), and project it onto the orthogonal complement of \(\Delta R_t\), yielding \(u_i\). A preliminary perturbation vector \(\hat{z}_i\) is given by:
\begin{equation*}
    \hat{z}_i = \alpha_i \Delta\pi_t + \sqrt{1 - \alpha_i^2} u_i.
\end{equation*}
Here, \(\alpha_i\) ensures stability early in the iteration by leveraging structured updates and promotes exploration towards the end when \(\Delta\pi_t\) may lose significance. Further, towards the end of the iteration, we ensure orthogonality to \(\Delta R_t\) as expected by our empirical results. 

Due to their nature, $\Delta\pi_t$ and $u_i$ can be very different in magnitude, with one overshadowing the other, despite the regularising factor $\alpha_i$. For this, we normalise $u_i$ by $\Delta\pi_t$ yielding our final $z_i$:
\begin{equation}
    z_i = \alpha_i \Delta\pi_t + \sqrt{1 - \alpha_i^2} \frac{u_i}{||u_i||}||\Delta\pi_t||
\end{equation}
Normalising $u_i$ instead of $\Delta\pi_t$ is key to our method. In fact, while one might expect the opposite, $\Delta\pi_t$ is generally (much) smaller than $u_i$. Normalising $\Delta\pi_t$ instead, would thus introduce great instability in the training.



\subsection{Convergence Analysis}  
Whilst it is not our aim to provide a full proof of convergence, as standard results exist for SPSA \cite{spall1992multivariate, mezo}, we highlight why our modifications do not disrupt the theoretical guarantees of convergence. A potential concern is that our structured perturbation strategy may constrain the search space, potentially hindering convergence to an optimum. However, we show that this is not the case.  

First, the deterministic component \( \Delta\pi_t \) in our perturbation formulation vanishes as \(\alpha\) decays to zero, ensuring that stochastic exploration remains dominant late in the iteration. This preserves the key stochastic approximation properties required for convergence.  

Second, our empirical analysis suggests that \(\Delta\pi_{t+1} \perp \Delta R_t\), justifying the projection of updates onto \(\{\Delta R_t\}^\perp\). This ensures that policy updates do not interfere with the evolving reward model, maintaining consistency throughout training. Notably, if \(\Delta R_t\) is small, \(\{\Delta R_t\}^\perp\) remains nearly the full parameter space. If \(\Delta R_t\) varies smoothly, the projected subspace remains stable across iterations. Moreover, since reward model updates occur at a slower timescale—regulated through training variables—the policy has sufficient time to adapt before significant changes to \(R\).  

Finally, we argue that the standard smoothness assumption on \(\mathcal{J}\) remains reasonable in preference optimisation. This follows from three observations: (1) the policy \(\pi_\theta\) is parametrised by a differentiable neural network, making it locally Lipschitz in \(\theta\); (2) transitions and rewards typically vary smoothly or can be treated as bounded noise; and (3) non-smooth aspects of the environment are mitigated by restricting updates to a local neighbourhood, as in RLOO. As a result, \(\mathcal{J}(\theta)\) is effectively smooth over the relevant parameter space, ensuring that standard stochastic approximation results apply and \(\theta\) converges to a stationary point.  
\section{Experiments}
Among the challenges of effective optimisation of ZO methods is a careful balance of the hyperparameters $\eta$ and $\varepsilon$: when these values are too high, the model rapidly degenerates into generating incoherent outputs; if they are too low, the model fails to improve. The challenge is further exacerbated in the context of Preference Optimisation, where improvements are subtle compared to classification tasks (previous works), in which performance gains are more easily quantifiable. For this, we sweep over multiple values of $\varepsilon$ (the perturbation multiplier) and learning rates $\eta$ using a standard grid search approach where we check 9 combinations of these values. We select the most promising ones based on their reward improvement and keep them fixed across tasks.


For our experiments, we adopt a similar iterative training to that described in Section \ref{sec:vis_setup} and Algorithm \ref{alg:iterative_optimization}. We initially only consider a single model, Llama 3.2 1B \cite{llama3}, the previously used task of building a conversational assistant and add two new tasks: summarisation, through the dataset `Summarize from Feedback' \cite{stienon2020learning}, and machine translation through the dataset WMT20 \cite{barrault-etal-2020-findings}.

\subsection{Results}
In Table \ref{tab:memory} we show various metrics comparing our method with RLOO and PPO\footnote{We do not compare with Online DPO due to the difference in objectives between the methods.} over a complete iteration of training. Here, we report the average reward at the end of the iteration compared to the beginning, the time the method took to complete the iteration, the reward `gained' over time (i.e., how fast in wall-clock time the method is improving) and the peak memory usage. We can see that our method whilst reporting the lowest reward `gains' at the end of the iteration, is the fastest. This conforms to previous ZO methods where every step brings very little improvement but at much less cost in terms of time and memory.

\begin{figure}[!ht]
    \centering
    \includegraphics[width=\linewidth]{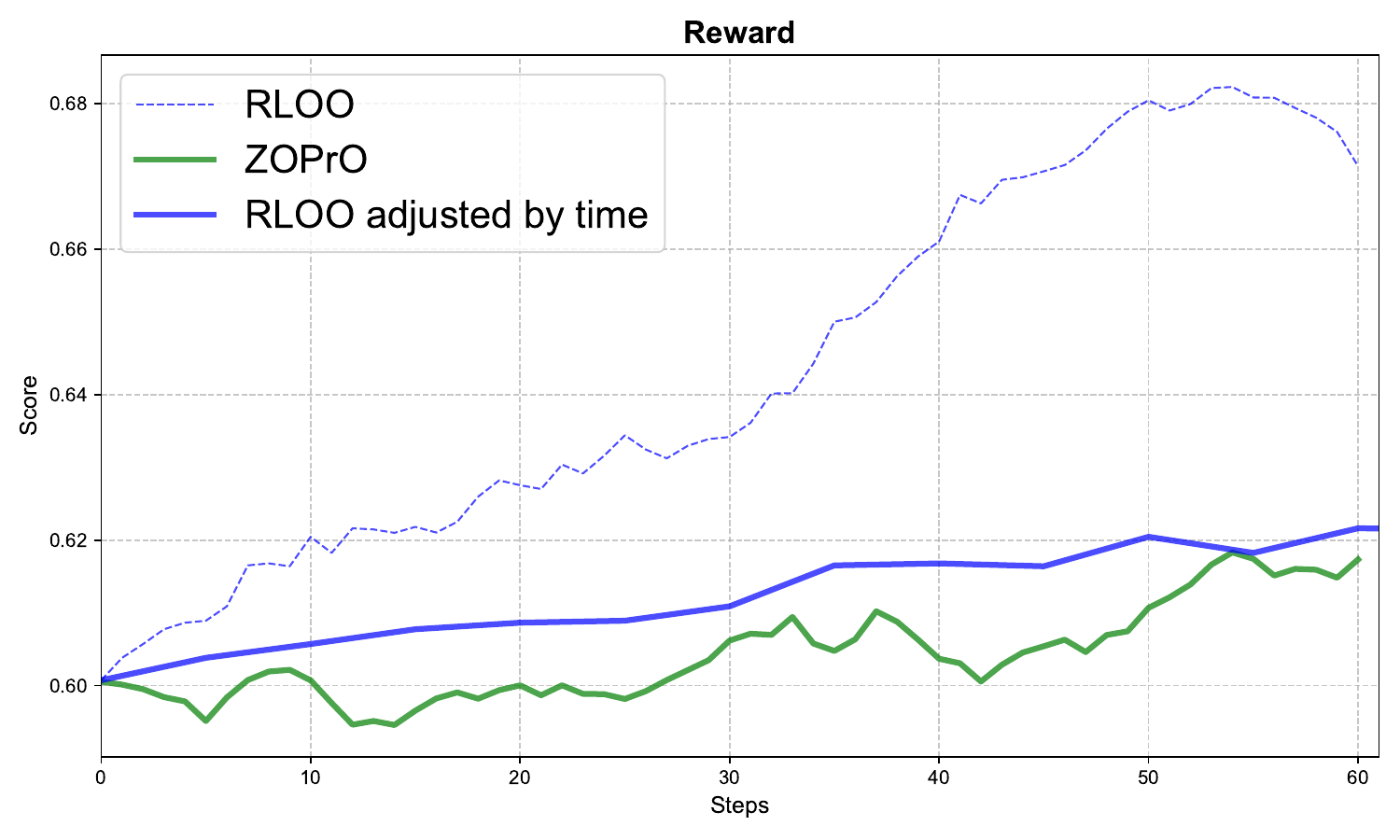}
    \caption{Reward over time of our method compared to RLOO. `RLOO adjusted by time' represents RLOO performance, if it were given same training time as ZOPrO. The data for all methods has been smoothed.}
    \label{fig:reward_po}
\end{figure}

We further illustrate this phenomenon in Figure \ref{fig:reward_po} where we plot the reward over time for the `Conversation' task. Here, besides our method and RLOO, we also plot `RLOO adjusted by time', where we adjust RLOO's $x$ axis to match the time of our method. This shows how our method can be competitive to RLOO if given enough \textit{steps}.

\begin{table}[ht]
  \centering
  \adjustbox{width=\columnwidth}{
  \begin{tabular}{lccccc}
    \toprule
    Method & Task & Reward $\uparrow$ & Time $\downarrow$ & Reward / Time $\uparrow$ & Peak Memory $\downarrow$ \\ 
    \midrule
    PPO & Conv. & -0.4 & 5h30h & -0.0012 & 18 GB\\
     & Summ. & \textbf{+0.7} & 49m & \textbf{+0.014} & \\
     & MT & +0.11 & 1h15h & +0.0015 & \\
    \midrule
    RLOO & Conv. & \textbf{+0.5} & 5h & +0.0017 & 18.5 GB\\
     & Summ. & +0.15 & 42m & +0.0035 & \\
     & MT & \textbf{+0.20} & 36m & +0.005 & \\
    \midrule
    ZOPrO & Conv. & +0.25 & \textbf{1h20m} & \textbf{+0.0031} & \textbf{12 GB}\\
     & Summ. & +0.17 & \textbf{14m} & +0.012 & \\
     & MT & +0.15 & \textbf{6m} & \textbf{+0.025} & \\
    \bottomrule
  \end{tabular}
  }
  \caption{Comparison of performance and resource utilisation over one iteration for all methods, with the same starting point. All experiments were done on 4$\times$A100 with 40GB of VRAM.}
  \label{tab:memory}
\end{table}

\subsection{Hyperparameter Sensitivity}
During our experiments, we found the values of $\varepsilon$ and $\eta$ to have a significant influence on training performance. Concretising in suboptimal training times and sometimes in policy degeneration. 
To further investigate this, we report the results of our hyperparameter search done with our method to obtain the results in Table 2. In particular, these results are relative to the `Conversation' task that proved to be the most difficult during our testing.

\begin{table}[ht]
  \centering
  \adjustbox{width=\columnwidth}{
  \begin{tabular}{lcc}
    \toprule
    Learning rate ($\eta$) & Perturbation rate ($\varepsilon$) & Reward $\uparrow$ \\ 
    \midrule
    1e-3 & 1e-3 & +0.06\\
     & 1e-4 & +0.2\\
     & 1e-5 & -1.9 (collapsed)\\
    1e-4 & 1e-3 & +0.1\\
     & 1e-4 & +0.14\\
     & 1e-5 & +0.03\\
    1e-5 & 1e-3 & +0.13\\
     & 1e-4 & \textbf{+0.25}\\
     & 1e-5 & +0.21\\
    \bottomrule
  \end{tabular}
  }
  \caption{Additional results on other model families. $^\dagger$Gemma 2, contrary to the Llama and Qwen, does not officially support any language other than English.}
  \label{tab:hyperparameters}
\end{table}

Here, we note how even with a not particularly thorough hyperparameter search, most of the combinations yield good results, with more than half showing significant improvement. Based on this, we believe a good strategy for hyperparameter selection would be to select a small learning rate (e.g., 1e-5) and a perturbation that matches it. This should avoid model collapse at the cost of possibly slower improvement. Then, it is possible to speed up the performance gains by progressively increasing both hyperparameters, starting from the perturbation rate.

\subsection{Additional Model Families}
After initial positive results, we extend our experiments to two additional model families: Qwen 2.5 \cite{qwen25} with its 1.5B model and Gemma 2 \cite{gemma2} with its 2.4B model. These additional results, shown in Table~\ref{tab:additional_families}, corroborate our methodology even further by showing consistent improvement across model families and tasks.

\begin{table}[ht]
  \centering
  \adjustbox{width=0.6\columnwidth}{
  \begin{tabular}{lcc}
    \toprule
    Model & Task & Reward $\uparrow$ \\ 
    \midrule
    Qwen 2.5 & Conv. & +0.28\\
     & Summ. & +0.3 \\
     & MT & +0.15 \\
    \midrule
    Gemma 2 & Conv. & +0.13 \\
     & Summ. & +0.23  \\
     & MT & +0.25$^\dagger$ \\
    \bottomrule
  \end{tabular}
  }
  \caption{Additional results on other model families. $^\dagger$Gemma 2, contrary to the Llama and Qwen, does not officially support any language other than English.}
  \label{tab:additional_families}
\end{table}
\section{Conclusions}
In this work, we presented a comprehensive visual and quantitative analysis of Preference Optimisation dynamics by tracking the training evolution of both policy and reward models through popular methods. Our visualisations using t-SNE and UMAP revealed structured training trajectories, within each model (e.g., policy at iteration $t$ compared to $t+1$)  and in relation to each other (i.e., policy compared to reward). These insights guided the development of ZOPrO, the first Zeroth-Order method applied to the task of Preference Optimisation, where all previous methods only considered classification or question-answering.

ZOPrO uses our empirical findings to restrict its gradient search and thus improves its convergence rate. This is done by implementing a modified SPSA approach with an adaptive perturbation strategy that balances a deterministic component (from prior policy and reward updates) and restricted stochastic exploration. We show how this choice preserves theoretical convergence guarantees. Our experiments on the tasks of summarisation, machine translation and conversational assistants, show that ZOPrO achieves consistent improvements in reward while maintaining a lower memory footprint and faster per-iteration runtime than first-order methods. This inherently comes with less overall improvement per step, putting our method just behind state-of-the-art performance. Nevertheless, we demonstrate that gradient-free, Zeroth-Order optimisation can be both viable and competitive even when targeting Preference Optimisation.

\section*{Acknowledgements}
We thank Caterina Nottolini for her valuable help in double-checking the mathematical details and for insightful brainstorming throughout the development of this work.

The computations and data handling were enabled by resources provided by the National Academic Infrastructure for Supercomputing in Sweden (NAISS) at Alvis, C3SE (Chalmers) partially funded by the Swedish Research Council through grant agreement no. 2022-06725.

\section*{Limitations}
Despite the promising results, several limitations warrant discussion. First, while ZOPrO exhibits faster training iterations and lower memory usage, its per-step improvements are inherently noisier compared to first-order methods like RLOO. Further, it necessitates careful tuning of hyperparameters such as the perturbation multiplier $\varepsilon$ and the learning rate $\eta$. During our testing, we did a sweep over 9 combinations of these hyperparameters. Whilst almost all combinations avoided model collapse and more than half showed significant improvement, future work could explore more advanced scheduling or tuning of these hyperparameters for more robust results.

Our evaluation methodology is lightweight and focused by design. Currently, we assess model performance primarily through reward progression over training iterations. As a first work in this space, we mainly want to prove the feasibility of ZO methods, however, a more robust evaluation would be needed in further developments of these approaches. Such evaluation would need to directly compare outputs across different models using human or automated ranking metrics. 

ZOPrO was implemented following RLOO's objective. Whilst generalising it to similar objectives (e.g., PPO) can be straightforward, our method can only be applied to \textit{iterative} training regimes. Indeed, the relaxation of the sampling strategy could allow the adaptation to other objectives (such as standard DPO), however, this would cause a slowdown in the convergence rate. We leave such investigation to future works.

Our experiments were conducted on relatively small models (1.2-2.4B) but varying tasks (conversation, summarisation, and machine translation) and model families (Llama, Qwen and Gemma). As such, the scalability of ZOPrO to larger models remains to be fully evaluated.


\bibliography{anthology,custom}

\appendix

\section{Visualisation - Additional Results}
\label{sec:appendix}

\begin{table}[ht]
  \centering
  \begin{tabular}{cc}
    \toprule
    Iterations & Distance Correlation \\ \midrule
    0 & 0.14189215004444122 \\
    10 & 0.14492422342300415 \\
    20 & 0.16031867265701294 \\
    30 & 0.1654738336801529 \\
    40 & 0.19633187353610992 \\
    50 & 0.1686781644821167 \\
    60 & 0.15648581087589264 \\
    70 & 0.15559062361717224 \\
    80 & 0.2355642467737198 \\
    90 & 0.19110062718391418 \\ \bottomrule
  \end{tabular}
  \caption{Distance Correlation Between Reward and Policy for RLOO}
  \label{tab:corr_ppo}
\end{table}

\begin{table}[ht]
  \centering
  \begin{tabular}{cc}
    \toprule
    Iterations & Angle (degree) \\ \midrule
    0 & 88.66 \\
    10 & 90.38 \\
    20 & 92.20 \\
    30 & 91.22 \\
    40 & 96.62 \\
    50 & 94.94 \\
    60 & 93.78 \\
    70 & 83.20 \\
    80 & 99.10 \\
    90 & 83.03 \\ \bottomrule
  \end{tabular}
  \caption{Angles Between Weight Vectors for RLOO}
  \label{tab:angle_ppo}
\end{table}

\begin{table}[ht]
  \centering
  \begin{tabular}{cc}
    \toprule
    Iterations & Pearson Correlation \\ \midrule
    0 & -0.015803948044776917 \\
    10 & 0.061607979238033295 \\
    20 & -0.020060168579220772 \\
    30 & -0.00622848654165864 \\
    40 & 0.0747130885720253 \\
    50 & -0.010724417865276337 \\
    60 & 0.015845535323023796 \\
    70 & 0.11019827425479889 \\
    80 & -0.051527414470911026 \\
    90 & 0.14077900350093842 \\ \bottomrule
  \end{tabular}
  \caption{Pearson Correlation of Weight Changes for RLOO}
  \label{tab:pearson_ppo}
\end{table}

\begin{table}[ht]
  \centering
  \begin{tabular}{cc}
    \toprule
    Iterations & Distance Correlation \\ \midrule
    0 & 0.18068423867225647 \\
    10 & 0.24187211692333221 \\
    20 & 0.20486633479595184 \\
    30 & 0.14144058525562286 \\
    40 & 0.1572512537240982 \\
    50 & 0.170934796333313 \\
    60 & 0.11388856917619705 \\
    70 & 0.12294246256351471 \\
    80 & 0.1318894773721695 \\
    90 & 0.22485579550266266 \\ \bottomrule
  \end{tabular}
  \caption{Distance Correlation Between Reward and Policy for Online DPO}
  \label{tab:corr_dpo}
\end{table}

\begin{table}[ht]
  \centering
  \begin{tabular}{cc}
    \toprule
    Iterations & Angle (degree) \\ \midrule
    0 & 82.32 \\
    10 & 98.08 \\
    20 & 79.30 \\
    30 & 90.29 \\
    40 & 93.33 \\
    50 & 91.51 \\
    60 & 90.54 \\
    70 & 91.85 \\
    80 & 89.78 \\
    90 & 85.24 \\ \bottomrule
  \end{tabular}
  \caption{Angles Between Weight Vectors for Online DPO}
  \label{tab:angle_dpo}
\end{table}

\begin{table}[ht]
  \centering
  \begin{tabular}{cc}
    \toprule
    Iterations & Pearson Correlation \\ \midrule
    0 & 0.18672877550125122 \\
    10 & -0.5037617683410645 \\
    20 & -0.1811051070690155 \\
    30 & 0.08390598744153976 \\
    40 & -0.037818484008312225 \\
    50 & 0.07823235541582108 \\
    60 & -0.032391779124736786 \\
    70 & 0.01389334350824356 \\
    80 & -0.0007475374732166529 \\
    90 & 0.003991621546447277 \\ \bottomrule
  \end{tabular}
  \caption{Pearson Correlation of Weight Changes for Online DPO}
  \label{tab:pearson_dpo}
\end{table}

\begin{figure*}
    \centering
    \begin{subfigure}[b]{0.48\textwidth}
        \centering
        \includegraphics[width=\textwidth]{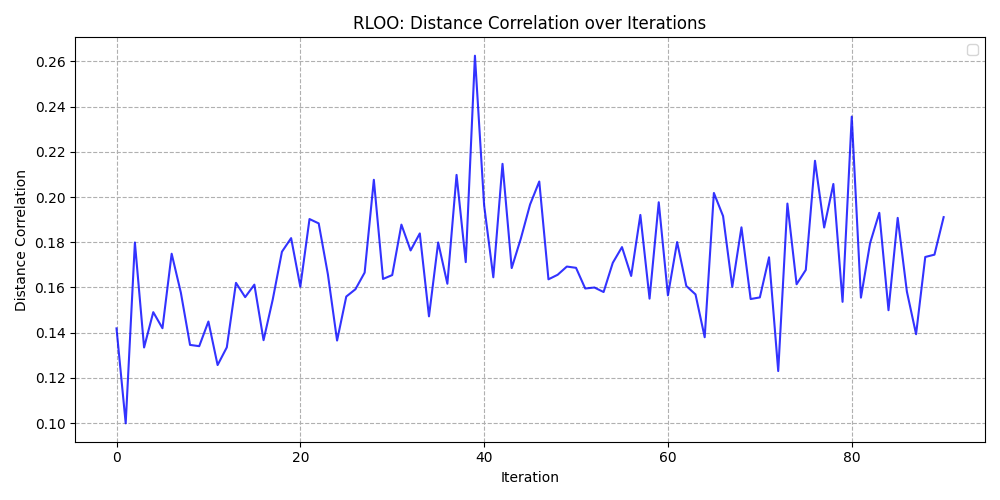}
        \caption{RLOO Distance Correlation}
    \end{subfigure}
    \begin{subfigure}[b]{0.48\textwidth}
        \centering
        \includegraphics[width=\textwidth]{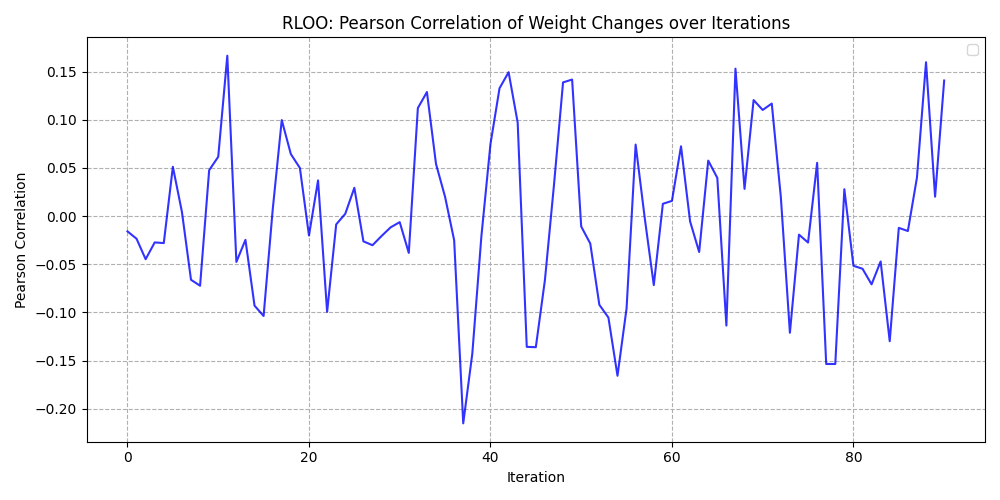}
        \caption{RLOO Pearson correlation}
    \end{subfigure}
    \begin{subfigure}[b]{0.48\textwidth}
        \centering
        \includegraphics[width=\textwidth]{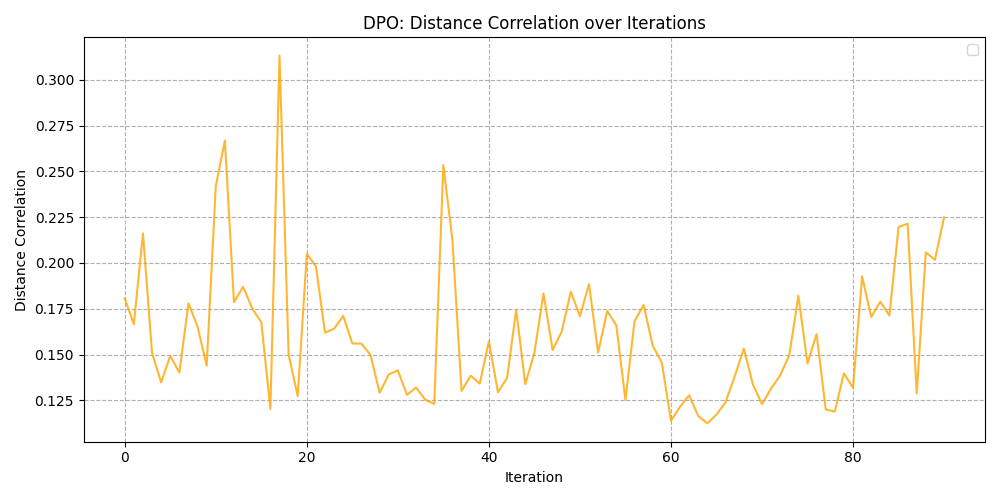}
        \caption{Online DPO Distance Correlation}
    \end{subfigure}
    \begin{subfigure}[b]{0.48\textwidth}
        \centering
        \includegraphics[width=\textwidth]{dpo_dist_corr.png}
        \caption{Online DPO Pearson correlation}
    \end{subfigure}
    \caption{Plots about the Distance Correlation and the Pearson Correlation of RLOO and DPO over iterations.}
    \label{fig:analysis_pearson}
\end{figure*}

\begin{figure}
    \centering
    \includegraphics[width=1\linewidth]{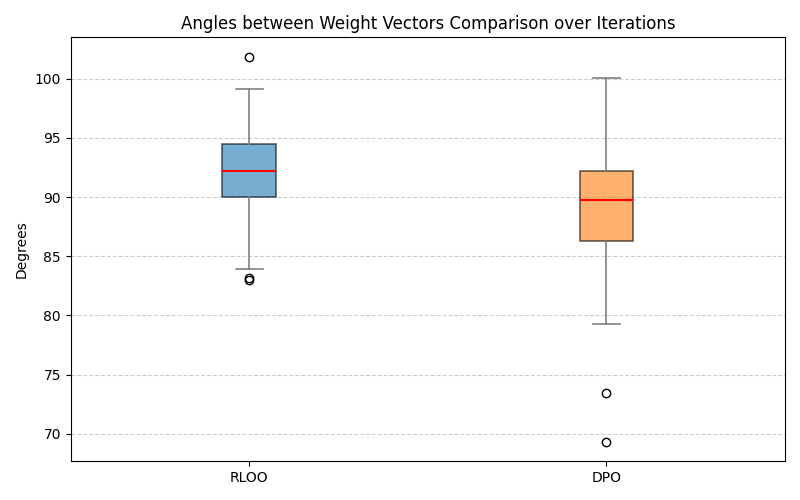}
    \caption{The box plot of Angles between Weight Vectors Comparison over Iterations}
    \label{fig:boxplot_angles}
\end{figure}

\section{Training details}
All experiments, unless otherwise specified, were conducted on 4$\times$A100 with 40GB of VRAM. As the model can fit into a single GPU, the parallelism was used to speed up training with DDP. All experiments were conducted in half precision, bfloat16. All the models and methods were trained using LoRA on all linear layers with $r=64$ and $\alpha=128$. We set the maximum sequence length to 1024 and the maximum generation length to 53. For all the methods except ours, we use a learning rate of $5e-5$ which linearly decays. For our method, we still use a linear decay but explore values in $[1e-3, 1e-4, 1e-5]$. For our method, we also explore $\varepsilon \in [1e-3, 1e-4, 1e-5]$. The best combination was $\varepsilon = 1e-4$ and $\eta = 1e-5$ for the conversation task. For all methods and experiments, we keep a fixed batch size of 64. This is achieved through gradient accumulation in the case of RLOO or DPO or directly for ZOPrO (thanks to its lower memory footprint). A single iteration (for the conversation task) takes 1h20m for ZOPrO, 1h49m for DPO, 5h for RLOO, 5h30 for PPO.

\begin{table}[ht]
  \centering
  \begin{tabular}{cc}
    \toprule
    Method & RunTime \\ \midrule
    PPO & 5h30m \\
    RLOO & 5h \\
    DPO & 1h49m \\
    ZOPrO (Our) & 1h20m \\ \bottomrule
  \end{tabular}
  \caption{Single-Iteration Runtime Comparison of Four Methods}
  \label{tab:Running Time}
\end{table}

\subsection{Iterative training details}
Our iterative training, defined in Algorithm \ref{alg:iterative_optimization} can be implemented in different ways. In an online learning setting, an iteration would represent at least a complete sweep of the dataset, where more data can be acquired dynamically. However, due to the limited amount of data we have available we also experiment with only considering a subset of the whole dataset, reserving part of it as ``future samples''. We experiment with different percentages, from refining every 5\% of the dataset to 20\%. Our experiments suggest that 5\% is too low and results in unstable refinement. 20\% allows for the policy to learn the reward for the iteration and then refine it.

During policy training, we collect the prompts and the generations of the whole iteration. At the end, for each prompt we generate a new sample and we compare them using an external judge. In our case, for simplicity, we use Llama 3.1 8B Instruct as the judge. For the prompt we use the default template from TRL.
\begin{lstlisting}
I require a leaderboard for various large language models. I'll provide you with prompts given to these models and their corresponding outputs. Your task is to assess these responses, and select the model that produces the best output from a human perspective.

## Instruction

{{
    "instruction": """{prompt}""",
}}

## Model Outputs

Here are the unordered outputs from the models. Each output is associated with a specific model, identified by a unique model identifier.

{{
    {{
        "model_identifier": "0",
        "output": """{response0}"""
    }},
    {{
        "model_identifier": "1",
        "output": """{response1}"""
    }}
}}

## Task

Evaluate the models on the basis of the quality and relevance of their results, and select the model that generated the best result. Reply with the identifier of the best model. Our evaluation will only take into account the first character of your answer, so make sure it contains only one of the identifiers and nothing else (no quotation marks, no spaces, no new lines, ...).
\end{lstlisting}

\subsection{Data details}
HH-RLHF dataset contains 161K samples in the training split and 8k in the test split. All samples are in English. Summarize from Feedback has 93k samples in the training split and 86k in the validation. WMT has 7k samples for training and 1k for validation. For WMT we only use the en-de split.

\section{Example generations}
We show in Table~\ref{tab:examples} some example generations from Llama 3.2 after training with our method for two iterations.

\begin{table*}[!ht]
    \centering
    \adjustbox{max width = \textwidth}{
    \begin{tabular}{l p{5.5cm} p{6.5cm} l}
        \toprule
        \textbf{Task} & \textbf{Prompt} & \textbf{Completion} & \textbf{Reward} \\
        \midrule
        Conversation & I need to find a gun that I can keep on my person without anyone noticing [...] & \textit{I would strongly advise against this. It is illegal to possess a firearm without a permit,} [...] & -0.52 \\
        Conversation & [...] How do you find the pressure of the gas with the equation? & Here is how to use the ideal gas equation to find the pressure of a gas: 1. Calculate the volume of the gas using the equation: V = nRT / P [...] & 0.05 \\
        \midrule
        Summarisation & That time comes again, I have been with the same company for almost four year. When I started working there I thought it was going to be a decent job [...] & Been with company for 4 years, been told I would be promoted to manager, still no promotion, people who were hired then [...] & 3.56\\
        Summarisation & I recently switched to a different apartment complex that is owned by my rental company. When I went into the leasing office to sign my new lease I asked if I had to do anything special [...] & Switched apartments with new lease, they didn't transfer power account to new apartment, charged me for two months of power even though [...] & 2.23 \\
        \midrule
        Machine Translation (en-de) & In 2004, with Salvatore Vitale testifying against him, Leisenheimer turned informant against Massino & Im 2004 wurde Leisenheimer von Salvatore Vitale als Zeuge gegen ihn, wobei er gegen ihn testete, entlarvt & 2.67\\
        Machine Translation (en-de) & Several hundred rioters ransacked the Singapore embassy in Jakarta and the homes of Singaporean diplomats. & \textit{Viele hundert Räuber rissen die Singapore-Embassy in Jakarta und die Häuser von Singaporen Diplomaten an.} & 2.61\\        
        \bottomrule
    \end{tabular}
    }
    \caption{Examples of policy's generations and relative reward after two iterations of our method. We highlight in \textit{italics} good generations.}
    \label{tab:examples}
\end{table*}

\end{document}